\documentclass{article}
\usepackage{amsfonts}
\usepackage{graphicx}
\usepackage{graphics}
\newcommand{\R}{\mathbb{R}}

\begin{document}

\title{Scale-Based Gaussian Coverings: Combining Intra and Inter 
Mixture Models in Image Segmentation}

\author{Fionn Murtagh$^{1,2,\star}$,  Pedro Contreras$^{2}$ and 
Jean-Luc Starck$^{3,4}$ \\
\\
$^{1}$ Science Foundation Ireland, Wilton Park House, \\ Wilton Place,
Dublin 2, Ireland \\
$^{2}$ Department of Computer Science, Royal Holloway \\ University of 
London, Egham TW20 0EX, England \\
$^{3}$ CEA-Saclay, DAPNIA/SEDI-SAP, Service d'Astrophysique, \\ 
91191 Gif sur Yvette, France \\
$^{4}$  Laboratoire AIM (UMR 7158), CEA/DSM-CNRS, \\
Universit\'e Paris Diderot \\
\\
$^\star$ Corresponding author: 
fmurtagh@acm.org}

\maketitle

\begin{abstract}
By a ``covering'' we mean a Gaussian mixture model fit to 
observed data.   Approximations of the Bayes factor can be availed of
to judge model fit to the data within a given Gaussian mixture model.  
Between families of Gaussian mixture models, we propose the R\'enyi
quadratic entropy as an excellent and tractable model comparison framework.
We exemplify this using the segmentation of an MRI image volume, 
based (1) on a direct Gaussian mixture model applied to the marginal
distribution function, and (2) Gaussian model fit through k-means 
applied to the 4D multivalued image volume furnished by the wavelet 
transform.  Visual preference for one model over another is not 
immediate.  The R\'enyi quadratic entropy allows us to show clearly 
that one of these modelings is superior to the other.  
\end{abstract}



\section{Introduction}

We begin with some terminology used.  Segments are contiguous clusters.  
In an imaging context, this means that clusters contain adjacent or 
contiguous pixels.  For typical 2D (two-dimensional) images, 
we may also consider the 1D (one-dimensional) marginal which 
provides an empirical estimate of the pixel (probability) density function
or PDF.
For 3D (three-dimensional) images, we can consider 2D marginals, based
on the voxels that constitute the 3D image volume, or also a 
1D overall marginal.  An image is 
representative of a signal.  More loosely a signal is just data, 
mostly here with neccessary sequence or adjacency relationships.  
Often we will use interchangeably the terms image, image volume if 
relevant, signal and data.  

The word ``model'' is used, in general, in many senses -- 
statistical \cite{mccul}, mathematical, physical models; mixture
model; linear model; noise model; neural network model; sparse
decomposition model; even, in different senses, 
data model.  In practice, firstly and 
foremostly 
for algorithmic tractability, models of whatever persuasion tend to be 
homogeneous.  In this article we wish to broaden the homogeneous 
mixture model framework in order to accommodate heterogeneity at 
least as relates to resolution scale.  
Our motivation is to have a rigorous model-based approach to 
data clustering or segmentation, that also and in addition 
encompasses resolution scale.

\begin{figure}
\begin{center}
\includegraphics[width=12cm]{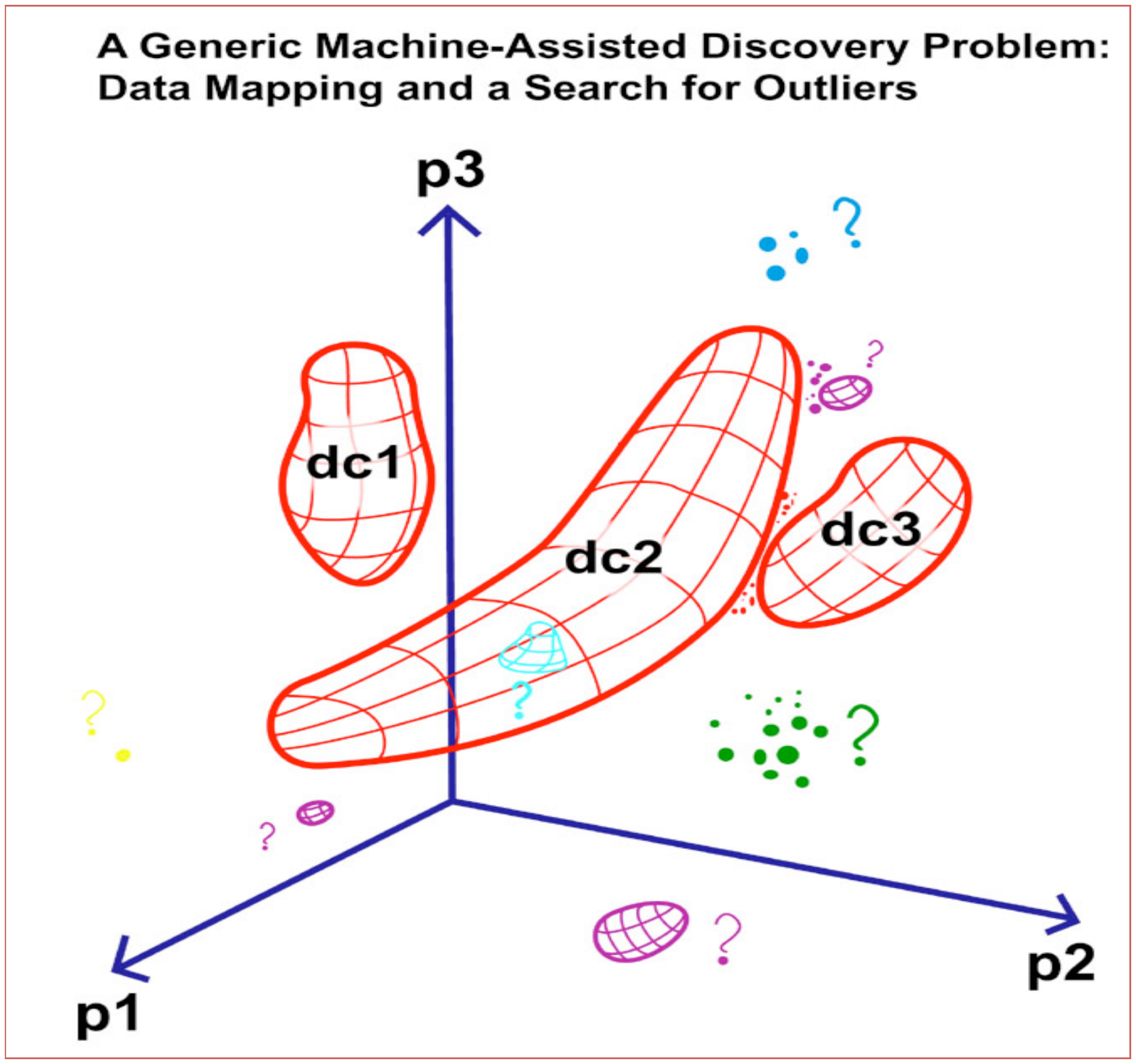}
\end{center}
\caption{Clusters of all morphologies are sought.  Figure courtesy 
of George Djorgovski, Caltech.}
\label{fig00}
\end{figure}

In Figure \ref{fig00} \cite{gdg}, the clustering task is portrayed in its
full generality.  One way to address it is to build up parametrized 
clusters, for example using a Gaussian mixture model (GMM), so that 
the cluster ``pods'' are approximated by the mixture made up of the 
cluster component ``peas'' (a viewpoint expressed by A.E.\ Raftery,
quoted in \cite{silk}).    

A step beyond a pure ``peas'' in a ``pod'' approach to 
clustering is a hierarchical approach.  
Application specialists often consider hierarchical 
algorithms as more versatile than their
partitional counterparts (for example, k-means or 
Gaussian mixture models)
since the latter tend to work well only on data 
sets having isotropic clusters \cite{nagy}.  So in 
\cite{ref12pt5}, we segmented astronomical images of different 
observing filters, that had first been matched such that they 
related to exactly the same fields of view and pixel resolution.
For the segmentation we used 
a Markov random field and Gaussian mixture model; followed by a
within-segment GMM clustering on the marginals.  Further within-cluster
marginal clustering could be continued if desired.  For model fit,
we used approximations of the Bayes factor: the pseudo-likelihood 
information criterion to start with, and for the marginal GMM work,
a Bayesian information criterion.  This hierarchical- or tree-based
approach is rigorous and we do not need to go beyond the Bayes factor
model evaluation framework.  The choice of segmentation methods used 
was due to the desire to use contiguity or adjacency information 
whenever possible, and when not possible to fall back on use of 
the marginal.  This mixture of segmentation models is a first 
example of what we want to appraise in this work.  

What now if we cannot (or cannot conveniently) match the images 
beforehand?   In that case, segments or clusters in one image will
not necessarily correspond to corresponding pixels in another image.
That is a second example of where we want to evaluate  different 
families of models.  

A third example of what we want to cater for in this work is 
the use of wavelet transforms to substitute for spatial modeling
(e.g.\ Markov random field modeling).
In this work one  point of departure is a Gaussian mixture model 
(GMM)
with model selection using the Bayes information criterion (BIC)
approximation to the Bayes factor.  We extend this to a new
hierarchical context.  We use GMMs on resolution scales of a 
wavelet transform.  The latter is used to provide resolution 
scale.  Between resolution scales we do not seek a strict
subset or embedding relationship over fitted Gaussians, but 
instead accept a lattice relation.  We focus in particular 
on the global quality of fit of this wavelet-transform based
Gaussian modeling.  We show that a suitable criterion of goodness
of fit for cross-model family evaluations 
is given by R\'enyi quadratic entropy.  

\subsection{Outline of the Article}

In section 2 
we review briefly how modeling, with Gaussian mixture modeling 
in mind, is mapped into information.  

In section 3 
we motivate Gaussian mixture modeling as a
general clustering approach.  


In section 4 
we introduce entropy and focus on the 
additivity property.  This property is importatant to us in 
the following context.  Since 
hierarchical cluster modeling, not well addressed or supported 
by Gaussian mixture modeling, is of practical importance 
we will seek 
to operationalize a wavelet transform approach to segmentation.  
The use of entropy in this context is discussed in section 5.

The fundamental role of Shannon entropy together with some other definitions 
of entropy in signal and noise modeling
is reviewed in section 6.  
Signal and noise modeling are
potentially usable for image segmentation.  

For the role of entropy in image segmentation, section 
2 
presented the state of the art relative to 
Gaussian mixture modeling; and section 6 
presented
the state of the art relative to (segmentation-relevant)
filtering.  

What if we have segmentations obtained through different modelings?
Section 7 
addresses this through the use of R\'enyi 
quadratic entropy.  Finally, section 8 
presents a case study.

\section{Minimum Description Length and Bayes Information Criterion}
\label{sectmdl}

For what we may term a homogenous modeling framework, the minimum description
length, MDL, associated with Shannon entropy \cite{rissanen}, 
will serve us well.  
However as we will now describe it does not cater for hierarchically 
embedded segments or clusters.  An example of where hierarchical 
embedding, or nested clusters, come into play can be found in 
\cite{ref12pt5}.

Following Hansen and Yu \cite{ref10}, we consider a model class, $\Theta$, 
and an instantiation of this involving parameters $\theta$ to be estimated,
yielding $\hat{\theta}$.  We have $\theta \in \R^k$ so the parameter space
is $k$-dimensional.  Our observation vectors, of dimension $m$, and of 
cardinality $n$, are defined as: $X = \{ x_i | 1 \leq i \leq n \}$.  A 
model, $M$, is defined as $f(X | \theta), \theta \in \Theta \subset \R^k,
X = \{ x_i | 1 \leq i \leq n \}, x_i \in \R^m, X \subset \R^m$.  
The maximum likelihood estimator (MLE) of $\theta$ is $\hat{\theta}$: 
$\hat{\theta} = \mbox{argmax}_\theta f(X | \theta)$.

Using Shannon information, the 
description length of $X$ based on a set of parameter values 
$\theta$ is: $ - \log f(X | \theta)$.  We need to transmit parameters also 
(as, for example, in vector quantization).  So overall code length is: 
$- \log f(X | \theta) + L(\theta)$.  
If the number of parameters is always the same, then $L(\theta)$ can be 
constant.  Minimizing $- \log f(X | \theta)$ over $\theta$ is the same 
as maximizing $f(X | \theta)$, so if $L(\theta)$ is constant, then 
MDL (minimum description length) is identical to maximum likelihood, ML.  


The MDL information content of the ML, or equally Bayesian maximum
a posteriori (MAP) estimate, is the code length of 
$ - \log f(X |  \hat{\theta}) + L(\hat{\theta})$.
First, we need to encode the $k$ coordinates of $\hat{\theta}$,
where $k$ is the (Euclidean) dimension of the parameter space.  
Using the uniform encoder for each dimension, the precision of coding is then 
$1/\sqrt{n}$ implying that 
the magnitude of the estimation error is  $1/\sqrt{n}$.  So the 
price to be 
paid for communicating  $\hat{\theta}$ is $k \cdot ( - \log 1/\sqrt{n} ) 
= \frac{k}{2} \log n$ nats \cite{ref10}.  
Going beyond the uniform coder is also possible with the same 
outcome.  

In summary, MDL with simple suppositions here 
(in other circumstances we could require more than two stages, and 
consider other coders) 
is the sum of code lengths for (i) encoding data using a given 
model; and (ii) transmitting the choice of model.  The outcome is 
minimal 
$- \log f(X | \hat{\theta}) + \frac{k}{2} \log n$.
  
In the Bayesian approach we assign a prior to each model class, and 
then we use the overall posterior to select the best model.  Schwarz's 
Bayesian Information Criterion (BIC), which approximates the Bayes factor of 
posterior ratios, takes the form of the same penalized 
likelihood, $ - \log f(X | \hat{\theta}) + \frac{k}{2} \log n$, where
$\hat{\theta} = $ ML or MAP estimate of $\theta$.  See \cite{fraley} for
case studies using BIC.

\section{Segmentation of Arbitrary Signal through a Gaussian Mixture Model}
\label{sect3}

Notwithstanding the fact that 
often signal is not Gaussian, cf.\ the illustration of Figure \ref{fig00}, 
we can  fit observational
data -- density $f$ with support in $m$-dimensional real space, $\R^m$ -- 
by Gaussians.  Consider the case of heavy tailed distributions.  

Heavy tailed probability 
distributions, examples of which include long memory or 
$1/f$ processes (appropriate for 
financial time series, telecommunications traffic flows,
etc.) can be modeled as a generalized Gaussian distribution (GGD,
also known as power exponential, $\alpha$-Gaussian distribution,
or generalized Laplacian distribution):

$$ f(x) = \frac{\beta}{2 \alpha \Gamma(1/\beta)} 
\exp{-( \mid x \mid / \alpha )^\beta}$$
where 

-- scale parameter, $\alpha$, represents the standard deviation, 

-- the gamma function, $\Gamma(a) = \int^{\infty}_0 x^{a-1} e^{-x} dx$, and

-- shape parameter, $\beta$, is the rate of exponential decay, $\beta > 0$.

A value of $\beta = 2$ gives us a Gaussian distribution.  A value of 
$ \beta = 1$ gives a double exponential or Laplace distribution.  For
$0 < \beta < 2$, the distribution is heavy tailed.  For $\beta > 2$, the 
distribution is light tailed.  

Heavy tailed noise can be modeled by a Gaussian mixture model with 
enough terms \cite{ref2}.  Similarly, in 
speech and audio processing, low-probability and large-valued noise 
events can be modeled as Gaussian components in the tail of the distribution. 
A fit of this fat tail distribution by a  Gaussian mixture model is 
commonly carried out \cite{ref20}.  As in 
Wang and Zhao \cite{ref20}, one can allow Gaussian component PDFs to recombine
to provide the clusters which are sought.  These authors also  found 
that using priors with heavy tails, rather than using standard Gaussian 
priors, gave more robust results.  But the benefit appears to be very 
small.

Gaussian mixture modeling of 
heavy tailed noise distributions, e.g.\
genuine signal and
flicker or pink noise constituting a heavy tail in the density,
is therefore feasible.
A solution is provided by a weighted sum of Gaussian densities often 
with decreasing weights corresponding to increasing variances.
Mixing proportions for small (tight) 
variance components are large (e.g., 0.15 to 0.3) whereas very large
variance components have small mixing proportions.  

\begin{figure}
\begin{center}
\includegraphics[width=14cm]{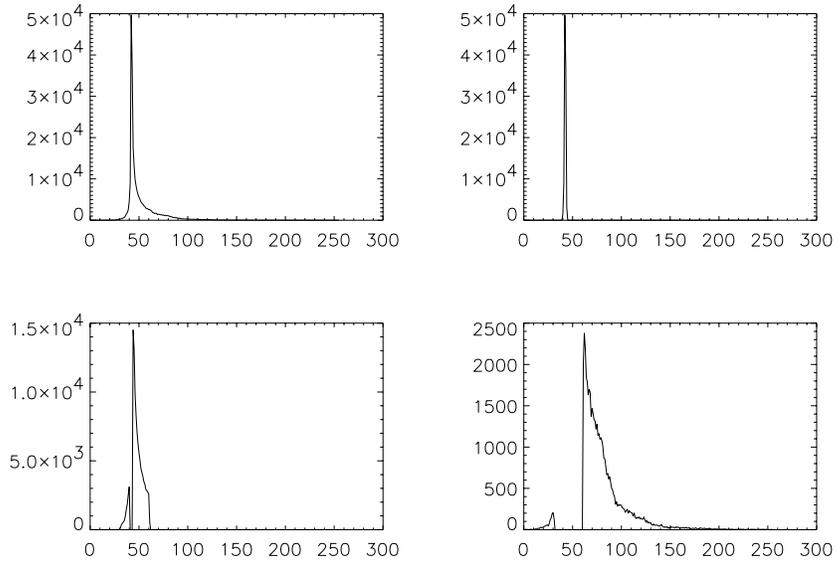}
\end{center}
\caption{Upper left: long-tailed histogram of marginal density of 
product of wavelet scales 4 and 5 of a $512 \times 512$ Lena image.  
Upper right, lower left, and 
lower right: histograms of 
classes 1, 2 and 3.  These panels exemplify a nested model.}
\label{xfig1}
\end{figure}

\begin{figure}
\begin{center}
\includegraphics[width=14cm]{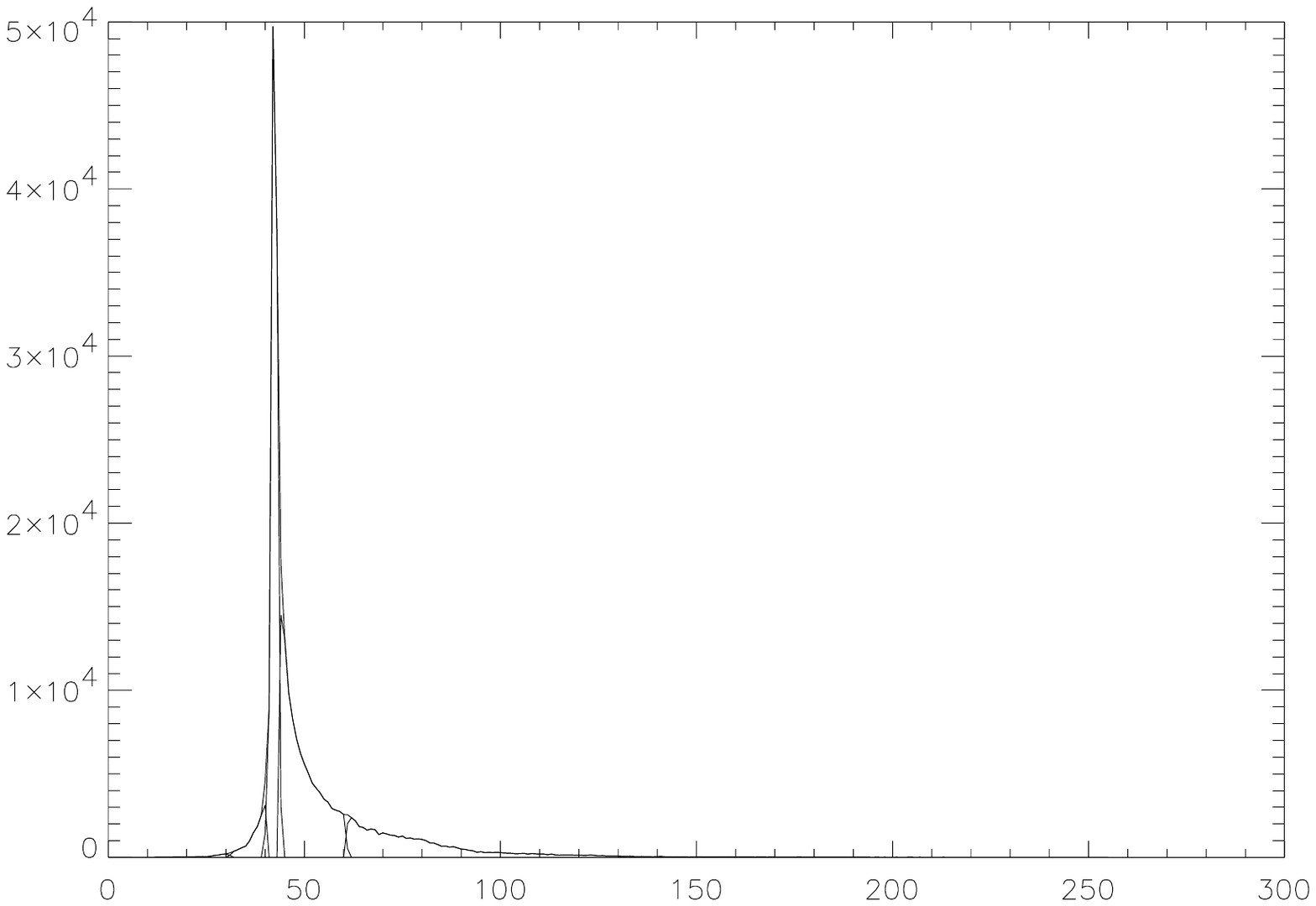}
\end{center}
\caption{Overplotting of the histograms presented in Figure \ref{xfig1}.
This shows how the classes reconstitute the original data.  The histogram
of the latter is the upper left one in Figure \ref{xfig1}.}
\label{xfig2}
\end{figure}

Figures \ref{xfig1} and \ref{xfig2} 
illustrate long-tailed behavior and show how 
marginal density Gaussian model fitting works in practice.
The ordinates give frequencies.  See further discussion in 
\cite{ref12,quant}.

\section{Additive Entropy}
\label{addent}

Background on entropy can be found e.g.\ in \cite{ref15}. 
Following Hartley's 1928 treatment of equiprobable events, Shannon 
in 1948 developed his theory around expectation.
In 1960 R\'enyi developed a recursive rather than linear estimation.
Various other forms of entropy are discussed in \cite{ref5}.

Consider density $f$ with support in $\R^m$.  Then: 

\begin{itemize}
\item Shannon entropy: $H_S = - \int f(x) \log f(x) dx$
\item 
R\'enyi entropy: $H_{R\alpha} = \frac{1}{1 - \alpha} \log \int f(x)^\alpha
dx $ for $\alpha > 0, \alpha \neq 1$.    
\end{itemize}

We have: $ \mbox{lim}_{\alpha \longrightarrow 1} H_{R\alpha} = H_S$.  
So $H_{R1} = H_S$.  
We also have: $H_{R\beta} \geq H_S \geq H_{R\gamma}$ for $ 0 < \beta < 1 $ and 
$1 < \gamma$ (see e.g.\ \cite{ref4}, section 3.3).  
When $\alpha = 2$, $H_{R2}$ is quadratic entropy.

Both Shannon and R\'enyi quadratic entropy are additive, a property 
which will be availed of by us below for example when we we define entropy for 
a linear transform, i.e.\ an additive, invertible decomposition.  

To show this, let us consider a system decomposed into independent 
events, $A$, $B$.

So $p(A B) $ (alternatively written: $p(A \& B)$ or $p(A + B)$) $ = 
p(A) p(B)$.  Shannon information is then $I_S^{AB} = - \log p(AB) 
= - \log p(A) - \log p(B)$, so the information of independent events is
additive.  Multiplying across by $p(AB)$, and taking $p(AB) = p(A) $ when 
considering only event $A$ and similarly for $B$, leads to additivity of 
Shannon entropy for independent events, $H_S^{AB} = H_S^A + H_S^B$.  

Similarly for R\'enyi quadratic entropy we use $p(AB) = p(A)p(B)$ and we 
have: $ - \log p^2(AB) = - 2 \log p(AB) = - 2 \log \left( p(A) p(B) \right)
= - 2 \log p(A) - 2 \log p(B) = - \log p^2(A) - \log p^2(B)$.  

\section{The Entropy of a Wavelet Transformed Signal}
\label{sect5}

The wavelet transform is a resolution-based decomposition -- hence
with an in-built spatial model: see e.g.\
\cite{jlsfmspringer,ref19pt5}.

A redundant wavelet transform is most appropriate, even if decimated
alternatives can be considered straightforwardly too.  This is because
segmentation, taking information into account at all available 
resolution scales, simply needs all available information.  A 
non-redundant (decimated, e.g., pyramidal) wavelet transform is 
most appropriate for compression objectives, but it can destroy 
through aliasing  potentially important faint features.  

If $f$ is the original signal, or images, then the following 
family of redundant wavelet transforms includes various discrete
transforms such as the isotropic, B$_3$ spline, \`a trous transform,
called the starlet wavelet transform in \cite{ref19pt5}.

\begin{equation}
f = \sum_{s = 1}^S w_s + w_{S+1}
\end{equation}
where: $w_{S+1}$ is the smooth continuum, not therefore wavelet coefficients; 
$w_s$ are wavelet coefficients at scale $s$.  Dimensions of $f, w_s, w_{S+1}$
are all identical.  

Nothing prevents us having a 
redundant Haar or, mutatis mutandis, redundant biorthogonal 9/7
wavelet transform (used in 
the JPEG-2000 compression standard).  As mentined above, our choice of 
starlet transform is due to no damage being done, through decimation,
to faint features in the image.  As a matched filter the starlet 
wavelet function is appropriate for many types of biological, 
astronomical and other images \cite{ref19pt5}.

Define the entropy, $H$, of the wavelet transformed signal as the 
sum of the entropies $H^s$ at the wavelet resolution levels, $s$:
\begin{equation}
H = \sum_{s = 1}^S H^s 
\end{equation}

Shannon and quadratic R\'enyi entropies are additive, as noted in 
section 4.  
For additivity, independence of
the summed components is required.  A redundant transform does 
not guarantee independence of resolution scales, $s = 1, 2, \dots , S$. 
However in practice we usually have approximate independence.  Our argument 
in favor of bypassing indepence of resolution scales is based on 
the practical and interpretation-related benefits of doing so.

Next we will review the Shannon entropy used in this context.
Then we will introduce a new application of the R\'enyi quadratic 
entropy, again in this wavelet transform context.  


\section{Entropy Based on a Wavelet Transform and a Noise Model}
\label{burg}

Image filtering allows, as a special case, thresholding and 
reading off segmented regions.  Such approaches have been used for 
very fast -- indeed one could say with justice, turbo-charged --
clustering.  See \cite{turbo1,turbo2}.

Noise models are particularly important in the physical sciences 
(cf.\ CCD, charge-coupled device, 
detectors) and the following approach was developed in \cite{ref19}.
Observed data $f$ in the physical sciences 
are generally corrupted by noise, which is often additive and which 
follows in many cases a Gaussian distribution, a Poisson distribution, or
a combination of both.  Other noise models may also be considered.
 Using Bayes' theorem to evaluate the probability distribution of the 
realization of the original signal $g$,
knowing the data $f$, we have
 
\begin{eqnarray}
 \mathrm{p}(g|f) = \frac{\mathrm{p}(f|g).\mathrm{p}(g)}{\mathrm{p}(f)}
\label{eqn_bayes}
\end{eqnarray}
$\mathrm{p}(f|g)$ is the conditional 
probability distribution of getting the data 
$f$ given an original signal $g$, i.e.\ it represents the distribution 
of the noise. It is given, in the case of uncorrelated Gaussian 
noise with variance $\sigma^2$, by:
\begin{eqnarray}
 \mathrm{p}(f|g) = \mathrm{exp} \left(
-\sum_{pixels} \frac{ (f-g)^2}{2{\sigma}^2} \right)
\label{eqn_proba}
\end{eqnarray}
The denominator in  equation (\ref{eqn_bayes}) is independent of $g$ and
 is considered as a constant (stationary noise). 
$\mathrm{p}(g)$ is the a priori distribution 
of the solution $g$. In the absence of any information on the solution 
$g$ except its positivity, a possible course of action 
is to derive the probability
of $g$ from its entropy, which is defined from information theory.

If we know the entropy $H$ of the solution (we describe below
different ways to calculate it), 
we derive its probability by
\begin{eqnarray}
\mathrm{p}(g) = \mathrm{exp}(- \alpha H(g))
\label{info_prop}
\end{eqnarray}

Given the data, the most probable image is obtained by maximizing
$\mathrm{p}(g|f)$.   This leads to algorithms for 
noise filtering and to deconvolution
\cite{jlsfmspringer}. 

We need a  probability density $p(g)$ of the data.
The Shannon entropy, $H_S$ \cite{ref18}, 
is the summing of the following for each pixel, 
\begin{eqnarray}
H_S(g) = - \sum_{k=1}^{N_b} p_k \log p_k
\end{eqnarray}
where  $X=\left\{g_1,.. g_n \right\}$ is an image 
containing integer values, $N_b$ is the number of possible values of 
a given pixel $g_k$ 
(256 for an 8-bit image), and the 
 $p_k$ values are derived from the histogram of $g$ as 
$p_k = \frac{m_k}{n}$, where $m_k$ is the number of occurrences 
in the histogram's $k$th bin. 

The trouble with this approach is that, because the number of occurrences is
finite, the estimate $p_k$ will be in error by an amount proportional
to $m_k^{-\frac{1}{2}}$ \cite{ref7}. The error becomes significant when
$m_k$ is small. Furthermore this kind of entropy definition is not
easy to use for signal restoration, because its gradient 
is not easy to compute.  For these reasons, other
entropy functions are generally used, including: 
\begin{itemize}
\item Burg \cite{ref3}:
\begin{eqnarray}
H_B(g) = -\sum_{k=1}^n \ln(g_k) 
\end{eqnarray}
\item Frieden \cite{ref6}:
\begin{eqnarray}
H_F(g) = -\sum_{k=1}^n g_k \ln(g_k)
\end{eqnarray}
\item Gull and Skilling \cite{ref9}:
\begin{eqnarray}
H_G(g) = \sum_{k=1}^n  g_k - M_k - g_k \ln \left( {g_k \over M_k} \right)
\end{eqnarray}
where $M$ is a given model, usually taken as a flat image
\end{itemize}
In all definitions 
$n$ is the number of pixels, and $k$ represents an index pixel.
For the three entropies above, unlike Shannon's entropy,
a signal has maximum information value when it is flat. 
The sign has been inverted
(see equation (\ref{info_prop})), to arrange for the best 
solution to be the smoothest.

Now consider the 
entropy of a signal as the sum of the information at each scale of its
wavelet transform, and the information of a wavelet 
coefficient is related to the probability of it being due to noise. Let
us look at how this definition holds up in practice.  
Denoting  $h$ the information relative to a single wavelet coefficient,
we define 
\begin{eqnarray}
H(X) = \sum_{j=1}^{l} \sum_{k=1}^{n_j}  h(w_{j,k}) 
\label{eqn_statwave}  
\end{eqnarray}
with the information of a wavelet coefficient, 
$h(w_{j,k})  = - \ln p(w_{j,k})$,
(Burg's definition rather than Shannon's).
 $l$ is the number of scales, and
$n_j$ is the number of samples in wavelet band (scale) $j$. 
For Gaussian noise, and recalling that wavelet coefficients at a given 
resolution scale are of zero mean, we get
\begin{eqnarray}
 h(w_{j,k}) =  \frac{w_{j,k}^2}{2 \sigma_j^2} + \mbox{constant}
\end{eqnarray}
where $\sigma_j$ is the noise at scale $j$.  When we use the 
information in a functional to be minimized (for filtering, 
deconvolution, thresholding, etc.), the constant term has no effect
and we can  omit it.
We see that the information is proportional
to the energy of the wavelet coefficients.
The larger the value of a normalized wavelet coefficient, 
then the lower will be 
its  probability of being noise, and the 
higher will
be the information furnished by this wavelet coefficient. 

In summary, 

\begin{itemize}
\item Entropy is closely related to energy, and as shown can be 
reduced to it, in the Gaussian context.
\item Using probability of wavelet coefficients is a very good way of 
addressing noise, but less good for non-trivial signal.
\item Entropy has been extended to take account of resolution scale.
\end{itemize}

In this section we have been concerned with the following view 
of the data: $f = g + \alpha + \epsilon$ where observed data $f$ is comprised 
of original data $g$, plus (possibly) background $\alpha$ (flat signal,
or stationary noise component), plus noise $\epsilon$.   
The problem of discovering signal from noisy, observed data is 
important and highly relevant in practice but it has taken us some
way from our goal of cluster or segmentation modeling of $f$ -- 
which could well have been cleaned and hence approximate well $g$ 
prior to our analysis.
  
An additional reason for discussing the work reported on in this 
section is the common processing platform provided by entropy.  

Often the entropy provides the optimization criterion used (see
\cite{ref8,ref15,jlsfmspringer}, and many other works besides). 
In keeping with entropy as having a key role in a common 
processing platform we instead want to use entropy for 
cross-model selection.  Note that it complements other criteria
used, e.g.\ ML, least squares, etc.  
We turn now towards a new way to define entropy for application 
across families of GMM 
analysis, wavelet transform based approaches, and other approaches 
besides, all with the aim of furnishing alternative segmentations.  

\section{Model-Based R\'enyi Quadratic Entropy}
\label{sect7}

Consider a mixture model: 
\begin{equation}
f(x) = \sum_{i = 1}^k \alpha_i f_i(x) \mbox{ with } 
\sum_{i = 1}^k \alpha_i  = 1
\end{equation}  
Here $f$ could correspond to one level of a mutiple resolution 
transformed signal.  
The number of mixture components is $k$.

Now take $f_i$ as Gaussian: 
\begin{equation}
f_i(x) = f_i(x | \mu, V) = 
\left((2 \pi)^{-\frac{m}{2}} 
| V_i |^{-\frac{1}{2}}\right) \exp \left( - \frac{1}{2} 
(x - \mu_i) V_i^{-1} (x - \mu_i)^t \right)
\end{equation}
where $x, \mu_i \in \R^m, V_i \in \R^{m \times m}$.  

Take 
\begin{equation}
V_i = \sigma_i^2 I
\label{eqnsi}
\end{equation} 
($I$ = identity) for simplicity of the basic 
components used in the model.  

A (i) parsimonious (ii) covering of these basic components can use a 
BIC approximation to the Bayes factor (see section 2) 
for selection of model, $k$.  

Each of the functions $f_i$ comprising the new basis for the observed 
density $f$ can be termed a {\em radial basis} \cite{ref11}.
A radial basis network, 
in this context, is an iterative EM-like fit optimization algorithm.  
An alternative view of parsimony is the view of a sparse basis, and 
model fitting is sparsification.  This theme of sparse or compressed
sampling is pursued in some depth in \cite{ref19pt5}. 

We have: 
\begin{equation}
\int^\infty_{-\infty} \alpha_i f_i (x | \mu_i, V_i) \ . \
\alpha_j f_j (x | \mu_j, V_j ) \ dx 
\end{equation}
\begin{equation}
= \alpha_i \alpha_j f_{ij} ( \mu_i - \mu_j, V_i + V_j) 
\end{equation}
See \cite{ref15} or \cite{ref8}.
Consider the case -- apropriate for us -- of only distinct clusters 
so that summing over $i, j$ we get:
\begin{equation} 
\sum_i \sum_j (1 - \delta_{ij}) \alpha_i \alpha_j f_{ij} 
(\mu_i - \mu_j , V_i + V_j) 
\end{equation}

Hence 
\begin{equation}
H_{R2} = - \log \int^\infty_{-\infty} f(x)^2 \ dx 
\end{equation}
can be written as:
\begin{equation}
- \log  \int^\infty_{-\infty} \alpha_i 
f_i(x | \mu_i, V_i) \ . \ \alpha_j
f_j(x | \mu_j, V_j) \ dx 
\end{equation}
\begin{equation}
 = - \log \sum_i \sum_j (1 - \delta_{ij}) 
\alpha_i \alpha_j 
f_{ij}( \mu_i - \mu_j , V_i + V_j)
\label{eqnx}
\end{equation}
\begin{equation}
 = - \log \sum_i \sum_j (1 - \delta_{ij}) 
f_{ij}( \mu_i - \mu_j , 2 \sigma^2 I)
\end{equation}
from restriction (\ref{eqnsi}) and also restricting the weights, 
$\alpha_i, \alpha_j = 1, \forall i \neq j$.  
The term we have obtained expresses interactions
beween pairs.  Function $f_{ij}$ is a Gaussian.  There are evident links 
here with Parzen kernels \cite{choi,jenssen} 
and clustering through mode detection 
(see e.g.\ \cite{rohlf}, and \cite{mur85} and references therein).

For segmentation we will simplify further expression (\ref{eqnx}) 
to take into account just the equiweighted segments reduced to their 
mean (cf.\ \cite{choi}).  

In line with how we defined mutiple resolution entropy in section 6,
we can also define the R\'enyi quadratic information of 
wavelet transformed data as follows: 
\begin{equation}
H_{R2} = \sum_{s = 1}^S H^s_{R2} 
\end{equation}

%

\section{Case Study}
\label{appl}

\subsection{Data Analysis System}

\begin{figure}
\begin{center}
\includegraphics[width=14cm]{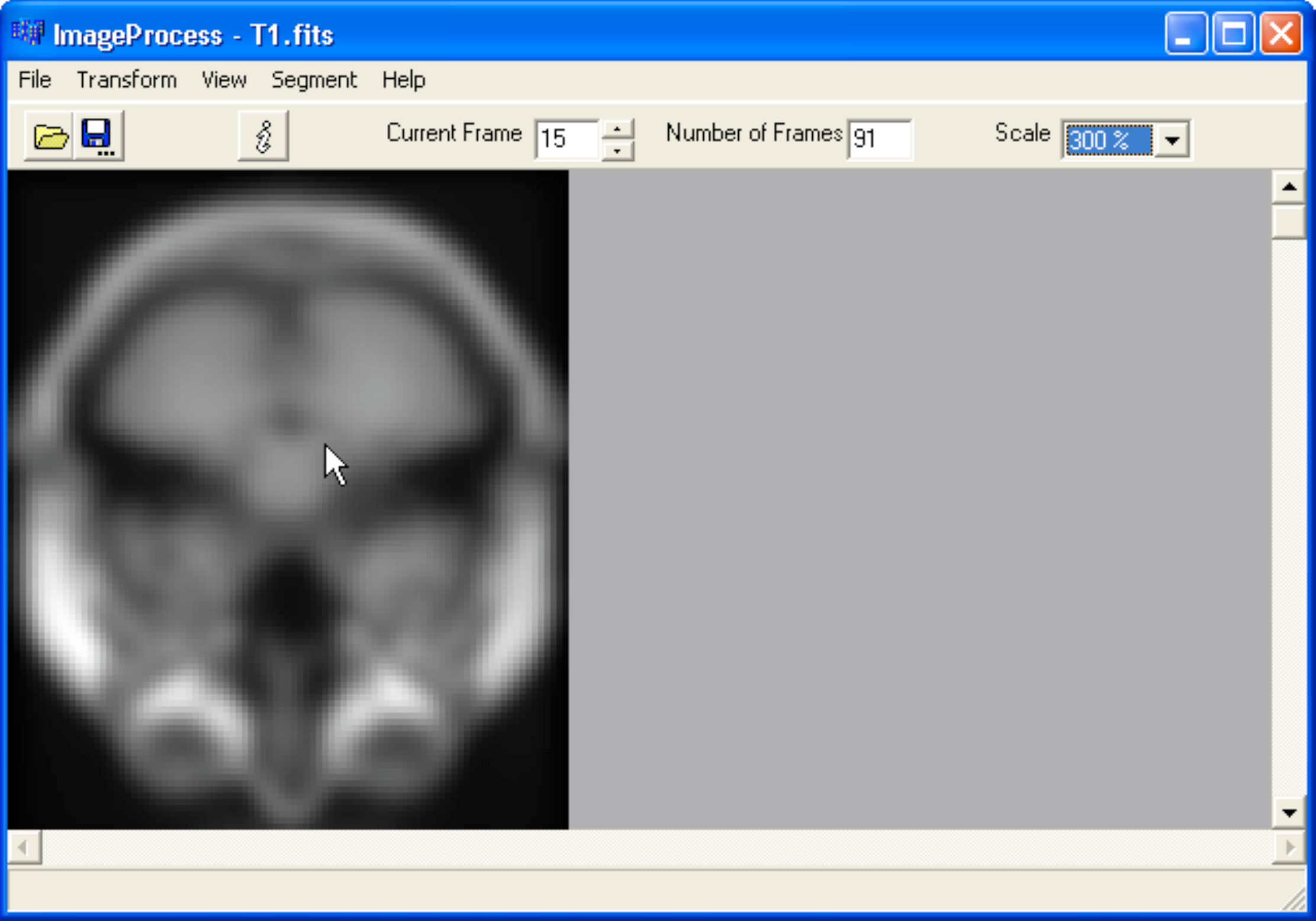}
\end{center}
\caption{Frame number 15 from an MRI brain image.}
\label{fig1}
\end{figure}

\begin{figure}
\begin{center}
\includegraphics[width=14cm]{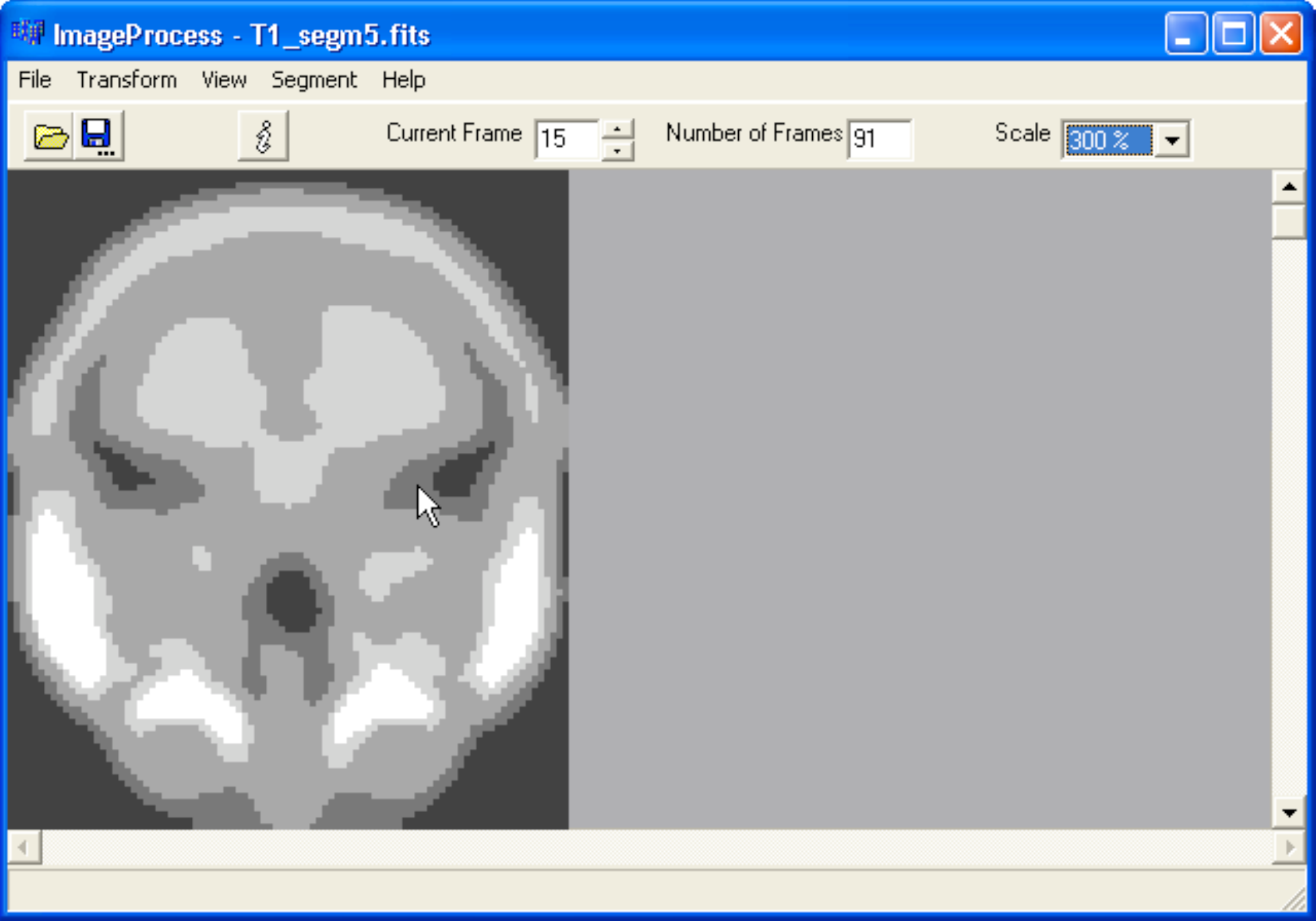}
\end{center}
\caption{Segmented frame number 15 from an MRI brain image.}
\label{fig2}
\end{figure}

In this work, we have used MRI (magnetic resonance imaging)  and 
PET (positron emission tomography) image data volumes, and 
(in a separate study) a 
data volume of galaxy positions derived from 3D cosmology data.
A 3D starlet or B$_3$ spline \`a trous wavelet transform is used with these 3D 
data volumes.  Figures \ref{fig1} and \ref{fig2} illustrate the system
that we built.  For 3D data volumes, we support the following formats:
FITS, ANALYZE (.img, .hdr), coordinate data (x, y, z), and DICOM; 
together with AVI video format.
For segmentation, we cater for marginal Gaussian mixture modeling, of a
3D image volume.  For multivalued 3D image volumes (hence 4D 
hypervolumes) we 
used Gaussian mixture modeling restricted to identity variances, and 
zero covariances, i.e.\ k-means.  Based on a marginal Gaussian mixture
model, BIC is used.  R\'enyi quadratic entropy is also supported.  
A wide range of options are available for presentation and display 
(traversing frames, saving to video, vantage point XY or YZ or XZ, 
zooming up to 800\%, histogram equalization by frame or image volume).  
The software, MR3D version 2, is available for download at 
www.multiresolution.tv.  The wavelet functionality 
requires a license to be activated, and currently the code has been 
written for PCs running Microsoft Windows only.   

\subsection{Segmentation Algorithms Used}

Consider T1, an aggregated representative brain, derived 
from MRI data.  It is of dimensions $91 \times 109 \times 91$.  See Figure
\ref{fig1}.  In the work described here as image format for the 
3D or 4D image volumes we used the FITS, Flexible Image Transport 
System, format.

The first model-based segmentation was carried out as follows.  

\begin{itemize}

\item We use ``Marginal Range'' in the ``Segmentation'' pull-down menu to 
decide, from the plot produced, that the BIC criterion suggests
that a 6 cluster solution is best.

\item Then we use ``Marginal'' with 6 clusters requested, again in the 
``Segmentation'' pull-down menu.  Save the output as:
T1\_segm\_marg6.fits.

\end{itemize}

Next an alternative model-based segmentation is carried out in 
wavelet space.  

\begin{itemize}

\item Investigate segmentation in wavelet space.  First carry 
out a wavelet transform.  The B$_3$ spline \`a trous wavelet transform
is used with 4 levels (i.e.\ 3 wavelet resolution scales).  The 
output produced is in files: 
T1\_1.fits,
T1\_2.fits,
T1\_3.fits,
T1\_4.fits.

\item Use the wavelet resolution scales as input to ``K-Means'', in the
``Segmentation'' pull-down menu.  Specify 6 clusters.  We used 6 clusters
because of the evidence suggested by BIC in the former modeling, and hence
for comparability between the two modelings.
Save the 
output as: T1\_segm\_kmean6.fits.

\end{itemize}

\subsection{Evaluation of Two Segmentations}

We have two segmentations.  The first is a segmentation found from
the voxel's marginal distribution function.  The second outcome is a 
segmentation found from the multivalued 3D (hence 4D) wavelet transform.  

Now we will assess T1\_segm\_marg6 versus T1\_segm\_kmean6.  If we
use BIC, using the T1 image and first one and then the second 
of these segmented images, we find essentially the same BIC 
value.  (The BIC values of the two segmentations
differ in about the 12th decimal place.)
Note though that the model used by BIC is the same as that used
for the marginal segmentation; but it is not the same as that 
used for k-means.  Therefore it is not fair to use BIC to 
assess across models, as opposed to its use within a family of 
the same model.  

Using R\'enyi quadratic entropy, in the ``Segmentation'' pull-down 
menu, we find 4.4671 for the marginal result, and 1.7559 for the 
k-means result.  

Given that parsimony is associated with small entropy here, this 
result points to the benefits of segmentation in the wavelet domain, 
i.e.\ the second of our two modeling outcomes.

\section{Conclusions}

We have shown that R\'enyi quadratic entropy provides an 
effective way to compare model families.  It bypasses the limits
of intra-family comparison, such as is offered by BIC.  

We have offered some preliminary experimental evidence too that 
direct unsupervised classification in 
wavelet transform space can be more effective than model-based
clustering of derived data.  Intended by the latter (``derived data'') 
are marginal distributions.  

Our innovative results are very efficient from computational and 
storage viewpoints.  The wavelet transform for a fixed number of resolution
scales is computationally linear in the cardinality of the input 
voxel set.  The pairwise interaction terms feeding the R\'enyi 
quadratic entropy are also efficient.  For both of these aspects 
of our work, iterative or other optimization is not called for.

\section*{Acknowledgements}

Dimitrios Zervas wrote the graphical user interface and 
 contributed to other parts of the software described in section 8.

\bibliographystyle{mdpi}
\makeatletter
\renewcommand\@biblabel[1]{#1. }
\makeatother

\end{document}